\documentclass[sigconf]{acmart}
\usepackage{booktabs} % For formal tables
\usepackage{bm}
\usepackage{xspace}

\newcommand{\fdist}{\ensuremath{{\text{d}}}\xspace}

\newcommand{\fkern}{\ensuremath{{\text{k}}}\xspace}

\newcommand{\vecx}{\ensuremath{\mathbf{x}}\xspace}
\newcommand{\vecy}{\ensuremath{\mathbf{y}}\xspace}

\newcommand{\haty}{\ensuremath{\hat{y}}\xspace}
\newcommand{\RR}{\ensuremath{\mathbb{R}}\xspace}
\newcommand{\Kmat}{\ensuremath{\mathbf{K}}\xspace}
\newcommand{\kvec}{\ensuremath{\mathbf{k}}\xspace}
\newcommand{\T}{\ensuremath{\text{T}}\xspace}

\begin{document}
\title{Prediction of neural network performance by phenotypic modeling}
%%% The submitted version for review should be ANONYMOUS

\author{Alexander Hagg}
%\orcid{1234-5678-9012}
\affiliation{%
  \institution{Bonn-Rhein-Sieg University of Applied Sciences}
  \city{Sankt Augustin} 
  \state{Germany} 
  \postcode{53757}
}
\email{alexander.hagg@h-brs.de}

\author{Martin Zaefferer}
%\orcid{1234-5678-9012}
\affiliation{%
  \institution{TH K\"oln, Institute for Data Science, Engineering, and Analytics}
  \streetaddress{Steinm\"ullerallee 2}
  \city{Gummersbach} 
  \state{Germany} 
  \postcode{51643}
}
\email{martin.zaefferer@th-koeln.de}

\author{J\"org Stork}
%\orcid{1234-5678-9012}
\affiliation{%
  \institution{TH K\"oln, Institute for Data Science, Engineering, and Analytics}
  \streetaddress{Steinm\"ullerallee 2}
  \city{Gummersbach} 
  \state{Germany} 
  \postcode{51643}
}
\email{joerg.stork@th-koeln.de}

\author{Adam Gaier}
\affiliation{%
  \institution{Bonn-Rhein-Sieg University of Applied Sciences}
  \city{Sankt Augustin} 
  \state{Germany} 
  \postcode{53757}\\
  \institution{Inria / CNRS / Universit\'e de Lorraine}  
  \city{Nancy}
  \country{France}
  \postcode{54000}
}

%% The default list of authors is too long for headers.
\renewcommand{\shortauthors}{Hagg, Zaefferer, Stork, Gaier}

\begin{abstract}
  Surrogate models are used to reduce the burden of expensive-to-evaluate objective functions in optimization. 
  By creating models which map genomes to objective values, these models can estimate the performance of unknown inputs, and so be used in place of expensive objective functions.
  Evolutionary techniques such as genetic programming or neuroevolution commonly alter the structure of the genome itself.
  A lack of consistency in the genotype is a fatal blow to data-driven modeling techniques: interpolation between points is impossible without a common input space.
  However, while the dimensionality of genotypes may differ across individuals, in many domains, such as controllers or classifiers, the dimensionality of the input and output remains constant.
  In this work we leverage this insight to embed differing neural networks into the same input space. 
  To judge the difference between the behavior of two neural networks, we give them both the same input sequence, and examine the difference in output. 
  This difference, the phenotypic distance, can then be used to situate these networks into a common input space, allowing us to produce surrogate models which can predict the performance of neural networks regardless of topology.
  In a robotic navigation task, we show that models trained using this phenotypic embedding perform as well or better as those trained on the weight values of a fixed topology neural network. We establish such phenotypic surrogate models as a promising and flexible approach which enables surrogate modeling even for representations that undergo structural changes. 
\end{abstract}

\begin{CCSXML}
  <ccs2012>
  <concept>
  <concept_id>10010147.10010257.10010293.10010075</concept_id>
  <concept_desc>Computing methodologies~Kernel methods</concept_desc>
  <concept_significance>500</concept_significance>
  </concept>
  <concept>
  <concept_id>10010147.10010257.10010293.10010294</concept_id>
  <concept_desc>Computing methodologies~Neural networks</concept_desc>
  <concept_significance>500</concept_significance>
  </concept>
  <concept>
  <concept_id>10010147.10010257.10010293.10011809.10011812</concept_id>
  <concept_desc>Computing methodologies~Genetic algorithms</concept_desc>
  <concept_significance>500</concept_significance>
  </concept>
  <concept>
  <concept_id>10010147.10010257</concept_id>
  <concept_desc>Computing methodologies~Machine learning</concept_desc>
  <concept_significance>300</concept_significance>
  </concept>
  </ccs2012>
\end{CCSXML}

\ccsdesc[500]{Computing methodologies~Kernel methods}
\ccsdesc[500]{Computing methodologies~Neural networks}
\ccsdesc[500]{Computing methodologies~Genetic algorithms}
\ccsdesc[300]{Computing methodologies~Machine learning}

\keywords{Surrogate Models, Neural Networks, Distance Metrics}

\copyrightyear{2019}
\acmYear{2019}
\setcopyright{acmlicensed}
\acmConference[GECCO '19 Companion]{Genetic and Evolutionary Computation Conference Companion}{July 13--17, 2019}{Prague, Czech Republic}
\acmBooktitle{Genetic and Evolutionary Computation Conference Companion (GECCO '19 Companion), July 13--17, 2019, Prague, Czech Republic}
\acmPrice{15.00}
\acmDOI{10.1145/3319619.3326815}
\acmISBN{978-1-4503-6748-6/19/07}

\maketitle

\section{Introduction}
% 一) What is the problem?
% 	- Expensive evaluation functions
% 	- design, robotics 

Optimization of real world engineering problems is a demanding task. Oftentimes, expensive simulations are needed to determine the quality of a solution. For example, to determine whether a car model produces low wind resistance, a numerical simulation of the airflow needs to be performed, which can take many hours or even days. In robotics control, we need to run physics-enabled simulations or run real world experiments. Iterative optimization requires many of these evaluations to reach a satisfactory solution.

% 二）How is it usually solved?
%     - Surrogate assisted optimization
% 	- non-parametric models
%   		- predictions based on known solutions
%   		- closer more similar

One of the most helpful techniques is to replace most evaluations with the predictions of a surrogate model~\cite{Jin2011,jin2018data}. The surrogate model is an efficient computational model that is trained with examples from the real objective function, but takes orders of magnitude less time to produce a prediction of the objective function's value for a certain candidate solution. Commonly used models such as Gaussian processes (also known as Kriging) or support vector machines~\cite{rasmussen2004,Jin2011}
are based on the similarity of candidate solutions.
Similarity-based surrogate models have been used in such varied domains as: shape optimization in fluid dynamics~\cite{ong2003evolutionary,daniels2018}, the discovery of new drugs~\cite{de2008active}, the placement of hospital trauma centers~\cite{wang2016data}, and even to the optimization of other machine learning methods~\cite{snoek2012,stork2017surrogate}.
To produce a prediction these models interpolate based on the distance of a candidate solution to known examples. 
They assume that the objective function is smooth: the closer a candidate is to a known example, the closer its function value will be to that of the example. 

% 三）Why that isn't sufficient?
% 	- Represnentation may not have constant input space 
% 		- GP, NEAT, architecture search
% 	- Behavior is an interaction with the environment
% 		- small difference in parameterization can result in large differences in behavior

\begin{figure}[h]
	\includegraphics{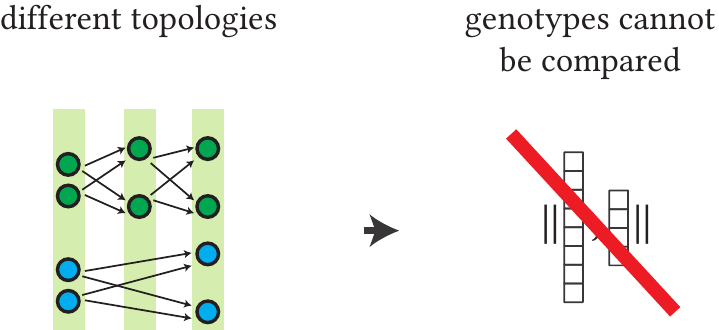}
	\caption{Two networks with different topologies cannot be compared based on their genotypes.}
	\label{fig:problem}
\end{figure}

A prerequisite for similarity-based surrogate models is that a distance metric is defined for the encoding of a solution. Surrogate models are therefore usually applied to solution representations that encode a fixed number of parameters. Recently, more complex encodings have been developed that do not have a constant input space. Prime examples of such encodings are compositional pattern producing networks (CPPN)~\cite{stanley2006exploiting}, that encode complex shapes or behaviors indirectly, neuroevolution~\cite{stanley2002}, in which the topology of neural networks can be evolved, or genetic programming~\cite{Koza1994}, which evolves the topology of graphs or trees representing computer code or mathematical equations. The non-uniform input space of these encodings frustrates typical ways of measuring distance as the dimensionality and even the meaning of these dimensions varies from one individual to the next (see Figure~\ref{fig:problem}).

A second problem arises when the quality of a solution depends on interaction with its environment. 
This behavior might vary greatly even if the parameterization of the encoding is changed only a small amount. 
If we would train a similarity-based model to predict the quality of such an encoding, a parameterization that is close to a training example would be assigned a similar fitness, although its actual fitness might be very different.  

% 四) What insight allows us to solve the problem?
% 	- Regardless of network composition input/output is the same
% 	- Input/output relation determines fitness
% 	- To measure the difference in the behavior of two networks we can feed them the same input sequence and measure the difference in the output

To enable surrogate-assisted optimization of these kinds of encodings, we investigate the idea of measuring distances not of the encoding, the genotype, but rather of the expression of the encoding, the phenotype. 
The phenotype may include morphological as well as behavioral aspects~\cite{dawkins1982extended}, and so can give us more information about how similar two individual solutions are than the genotype alone~\citep{Stor18c}. Our main insights are that (1) regardless of a network's internal composition, the size of the output in relation to the input is constant, and (2) the relation between input and output describes the behavior, and so is a useful proxy for similarity between networks. To measure the difference in the behavior of two networks we can give them the same input sequence and measure the difference in the output sequence using a standard metric like Euclidean distance.

% 五) How do we leverage this insight?
%     - The output difference can be used as a distance measure/kernel in a non-parametric model
%     - With such a model we can predict the performance of unevaluated networks, and so perform surrogate-assisted optimization regardless of representation

By using randomly selected, but fixed input sequences, we do not have to run an actual simulation to get the output sequence. Instead, we sample the input/output relation and use the ad hoc difference in the output sequences of two individuals to measure their distance. This distance measure can now be used to build a similarity-based surrogate model.

In this work, we evaluate whether we can model the phenotype of neural networks using a phenotypic distance metric and whether the models are competitive to those using a genotypic distance metric, which is based purely on the weights of the neural network. We qualify the results with a more in-depth analysis of the complexity of the phenotypic modeling problem, which shows that the intrinsic dimensionality of the phenotypic data is much lower than that of the genotypic data.

\section{Related Work}
Similar to phenotypic distances, semantic distances are used in Genetic Programming (GP). 
These semantic distances can be defined as a distance of the outputs of GP individuals, determined with the same measure that is used in the fitness function~\cite{moraglio2012geometric} . 
Semantic distances are applicable where the fitness function can be computed 
as a distance between the optimal target vector and the candidate outputs, 
such as in supervised classification or symbolic regression.
In these cases, the semantic distance has a fitness distance correlation of exactly one and can be utilized to construct specific mutation and crossover operators, rendering the problem uni-modal. 

Phenotypic distances have also been employed in a surrogate modeling context for GP.
Hildebrandt and Branke~\cite{Hildebrandt2014} suggested a phenotypic distance for dynamic job shop scheduling problems.
Their definition of phenotypic distance compares individuals according to the results of evolved dispatching rules on a small set of test situations.
Unlike semantic distances, their phenotypic distance is not identical to the measure used in the actual fitness function.
This is necessary in the context of surrogate modeling for expensive fitness functions. 
If the fitness function is expensive to compute, it would also be expensive to use the same evaluation to compute a distance between candidates. Such an approach would render the surrogate model itself expensive.

Zaefferer et al. \cite{Zaef18b} compare different genotypic and phenotypic distances for surrogate models in symbolic regression. 
Here, the underlying measure is not identical to that used in the fitness function. Specifically, the fitness function
considers fixed coefficients in the symbolic expression. These coefficients are otherwise optimized during an actual fitness evaluation, which may become costly.
In both of these cases, the phenotypic distance was reported to yield better results than genotypic distances~\cite{Hildebrandt2014,Zaef18b}.

Doncieux et al. \cite{doncieux2010behavioral} discuss the use of behavioral similarity in evolutionary robotics to employ a diversity measure for a multiobjective optimization approach. 
They compare different distances based on the states, outputs and trajectories given concrete robot tasks. 
They outline that using these behavioral distance as second objective in multi-objective optimization is able to enhance the overall performance. 

A first approach utilizing a surrogate model for evolving neural networks given complex control tasks was discussed by Gaier et al \cite{gaier2018data}. 
An evolutionary algorithm was combined with a surrogate model based on a hereditary distance, which is defined in the context of NeuroEvolution of Augmenting Topologies (NEAT) as \textit{compatibility distance}. 
The approach is able to significantly improve the evaluation efficiency. 
Stork et al. \cite{Stor18c} also investigated surrogate models
for neuroevolution. They examined simple classification tasks and compared a phenotypic distance measure to genotypic distances in surrogate-assisted Cartesian genetic programming. 
The use of a phenotypic distance was shown to be very promising in terms of evaluation efficiency.  

In this work we build on the ideas about using the output of network representations, and investigate whether sampling the phenotypes allows us to measure distances between networks. We evaluate whether we can use this distance metric to model the behavior of neural networks in a robot control task and predict their fitness.

\section{Methods}
\subsection{Kriging}\label{sec:Kriging}
To perform interpolation or regression on a given data set, Kriging models assume that the underlying data is sampled from a Gaussian process. For an in-depth introduction to Kriging and its application in model-based optimization, we refer to Forrester et al.~\cite{Forrester2008a}. We give only a rough overview, focusing on the issues relevant to this work.

Here, the training data of the model is denoted as a set of $n$ solutions $\bm{X}=\{ \vecx^{(i)} \}_{i=1\ldots n}$ in a $k$-dimensional search space.
The corresponding $n$ observations are denoted with $\vecy= \{ y^{(i)} \}_{i=1\ldots n}$.
For an unknown point in our search space, $\vecx^*$, Kriging intends to estimate the unknown function value $\haty(\vecx^*)$.
In its core, the model assumes that the observations at each location \vecx are correlated via a kernel function.
In this paper, we consider kernel functions of the following type:
\begin{equation}
\fkern(\vecx,\vecx') =\exp \left(\ -\theta \fdist(\vecx,\vecx') \right).
\end{equation}
This essentially expresses the correlation of two samples \vecx a \vecx', based on their distance $\fdist(\vecx,\vecx')$,
and a kernel parameter $\theta\in\RR^+$. Kernel parameters are usually determined by Maximum Likelihood Estimation (MLE),
that is, they are chosen such that the data has the maximum likelihood under the resulting model.
MLE usually involves a numerical optimization procedure~\cite{Forrester2008a}.
The distance measure $\fdist(\vecx,\vecx')$ can potentially be any measure, though not all ensure that the kernel
is positive semi-definite, a common requirement~\cite{Forrester2008a}.
In this work, we use the Manhattan distance, which is less affected by issues related to high-dimensional data~\cite{Aggarwal2001}, defined as: 

\begin{equation}
\fdist_\text{Man}(\vecx,\vecx')=\sum |x_i - x_i'|
\end{equation}

Rather than a single parameter $\theta$, a different $\theta$ can be used for each dimension $i$ of the input samples, enabling the model to estimate the influence of each individual dimension on the observed values. However, in the interest of simplicity and computational efficiency we opt for an isotropic kernel with a single $\theta$. 

Once the pairwise correlations between all training samples are collected in a matrix \Kmat, the Kriging predictor
can be specified with
\begin{equation}
\hat{y}(\vecx^*)=\hat{\mu}+\kvec^\T \Kmat^{-1} (\vecy-\bm{1}\hat{\mu}),
\end{equation}
where $\hat{\mu}$ is another model parameter (estimated by MLE),
$\kvec$ is the vector of correlations between
training samples $\bm{X}$ and the new sample $\vecx^*$,
and $\bm{1}$ is a vector of ones.
The error or uncertainty of the prediction can be estimated with
\vspace{-3pt} 
\begin{equation}
\vspace{-3pt} 
\label{eq:error}
\hat{s}^2(\vecx)= \hat{\sigma}^2  (1- \kvec^T \Kmat^{-1} \kvec^T ), 
\end{equation}
where $\hat{\sigma}^2$ is a further model parameter to be estimated by MLE. 

\subsection{Genotypic vs Phenotypic Distance}
The networks that we investigate in this work are results of optimization runs with fixed network topologies.
This allows us to evaluate and compare the efficiency of models based on both genotypic and phenotypic 
distance measures.
To define a genotypic distance we consider the vector of weights of the neural networks.
Let $\vec{w}_{\vecx} = \{w_1,w_2,...,w_j\}$ be the weight vector of length $j$ associated with a solution $\vecx$, then we can calculate
the genotypic distance by the related weights of two samples: $d(\vec{w},\vec{w}')$.

The disadvantage of genotypic distance measures is their lack of applicability when changing topologies are considered. 
If in these cases no clear concept to compare genotypic changes exist (as applied in \cite{gaier2018data}), the genotypic distance comparison is difficult, misleading and even destructive {\cite{Stor18c,doncieux2010behavioral}.
	The ability to compare non-uniform topologies makes phenotypic distances a valuable technique, especially in cases when typical distances are not a viable option.
	
	The phenotype displays the behavior of a neural network given a certain set of inputs.
	For example, in the case of neural networks used as controllers for robots the phenotype can be defined as the control commands 
	that are issued in response to different sensor inputs. We define a phenotypic distance as follows:
	Let $\vec{s}=\{s_1,s_2,...,s_k\}$ be the vector of inputs with length $k$, then $\vec{o}_{\vecx}=\{o_1,o_2,...,o_{k \times z}\}$ is the associated processed output vector, or phenotype,
	for a neural network $\vecx$ with length $k \times z$, where $z$ is the number of neural network output neurons. 
	The phenotypic distance is employed by calculating the difference in the outputs of two samples: $d(\vec{o},\vec{o}')$.
	Figure~\ref{fig:network} illustrates the sampling of phenotypes and Figure \ref{fig:distance} shows a comparison of both distances.
	
	The phenotypic distance is always task sensitive, i.e., a comparison of two samples $\vecx$ and $\vecx'$ requires the definition of an adequate input vector $\vec{s}$.
	In the context of model-based optimization, this input vector needs to fulfill two requirements:
	\begin{itemize}
		\item[a)] the input should be representative for the underlying task, i.e., in case of robot control it should follow the given sensor ranges and/or depict a trajectory of states present in the task. 
		\item[b)] the dimensionality of the phenotype needs to be considered, the length of the input vector for generating the phenotypes might significantly affect the modeling performance as well as the computation time for querying the networks. 
	\end{itemize}
	Given a carefully selected input vector, the phenotypic distance should be able to provide a clear impression of how
	the behaviors of two candidate networks compare to each other. 
	A possible disadvantage of our definition of a phenotypic distance is that depending on the underlying task, the real behavior cannot be defined by the output of the neural network controller alone. For example, a robot is further influenced by the structure of the environment and its own body. Two robots with different controllers and phenotypes, one that uses 4 legs for movement and the other that uses 3 legs, might behave the same if the 4th leg is disabled due to damage \cite{doncieux2010behavioral}.
	However, a representative set of samples of the input/output relationship should be descriptive enough to capture the behavioral differences and so allow the construction of surrogate models.
	
	\begin{figure}
		\includegraphics{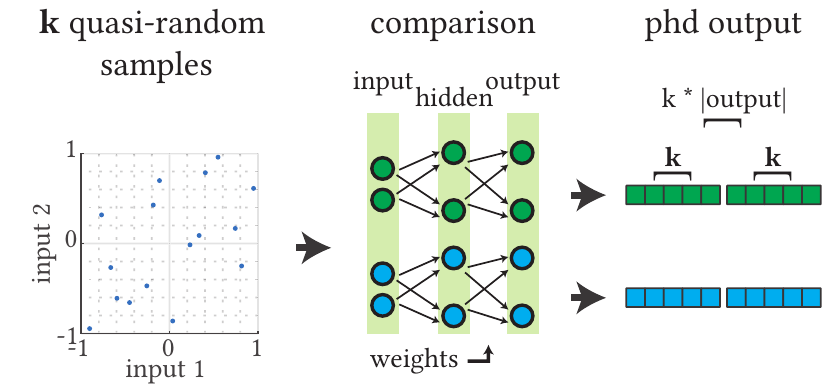}
		\caption{Sampling the phenotype to compare two individual networks.}
		\label{fig:network}
	\end{figure}
	
	\begin{figure}
		\includegraphics{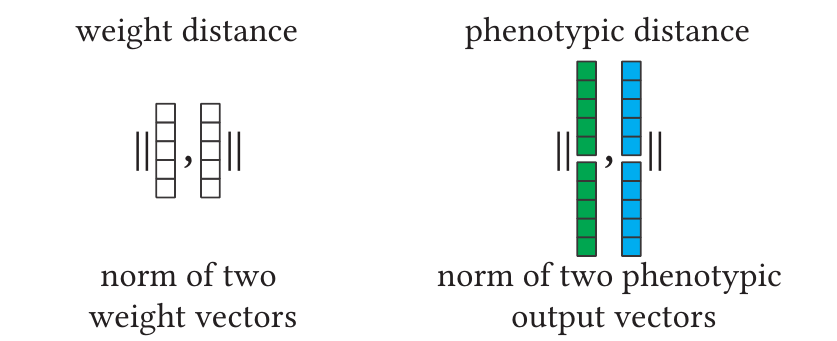}
		\caption{Weight models are based on weight vectors for fixed-topology networks. Phenotypic distance models are based on fixed-length sampled phenotypic output vectors for any-topology networks.  We use the L1 norm (Manhattan distance) for interpolative modeling.}
		%Norm of the distance used for interpolative modeling is either Manhattan or Euclidean.}
		\label{fig:distance}
	\end{figure}
	
	\section{Evaluation}
	\subsection{Experimental Setup}
	The goal of our experiments is two-fold.
	Firstly, to determine whether we can learn reasonable surrogate models based on a diverse set of phenotypic vectors. 
	Secondly, to compare these results to a genotypic model.
	Our experiments are constructed as follows, we:
	\begin{enumerate}
		\item Run model-free optimization algorithms that optimize the weights of fixed topology neural networks for robot control;
		\item Archive a selection of several hundred diverse neural networks from the results of these runs;
		\item Train different genotypic and phenotypic surrogate models on a subset of these networks;
		\item Test the performance of the surrogate models by predicting the performance of the remainder of the networks.
	\end{enumerate}
	
	\paragraph{Problem Setup: Maze and Robot}
	We design robot controllers for the multi-modal maze problem depicted in Figure~\ref{fig:maze}. The environment consists of multiple rings and openings (Figure~\ref{fig:maze}a). The robot begins in the center of the maze. Here we are not interested in typical case of finding the best solution to escape the maze. Instead, we seek to establish to what degree we can sample the behavior (or phenotype) of neural network controllers and then derive fitness from those behaviors. This problem is much more fundamental and difficult than predicting fitness alone. To produce this data set of as many different high performing behaviors as possible, we build up an archive out of robot controllers that reach every point in the maze in the shortest path possible (b). To force a diversity of ending positions, a grid-like diversity measure is defined (c). At the end of the optimization, every niche should contain a robot that was able to reach it using a short path. This way we can evaluate the distance measures over a diverse set of optimal behaviors.
	
	\begin{figure}[hbp]
		\includegraphics{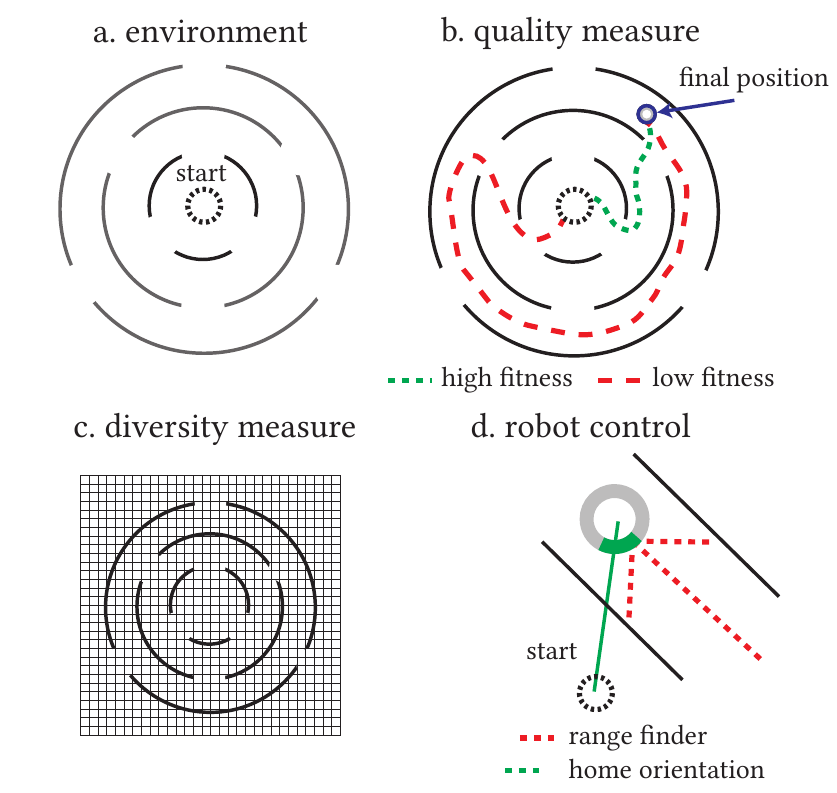}
		\caption{Evaluation takes place in a maze environment (a) with a robot starting in the center. The distance of the path of a robot to its final position defines its quality (b), whereby a diversity measure allows us to train robots to reach all cells in the map (c). Robots can sense the orientation quadrant of the start position and uses three range finders to perceive the distance to the nearest wall (d).}
		\label{fig:maze}
	\end{figure}

	Simple feed forward controllers (see Figure~\ref{fig:network}) consisting of either 2 or 5 hidden neurons are sought that traverse the maze. Evaluation is performed using the simulation that was created in~\cite{Mouret2011}. The robot is equipped with three laser sensors that are able to detect the distance to the nearest walls, and are set at 45 degree angles around the front (d).
	In addition, each robot has a home beacon that detects the direction of the robot's start position. 
	
	\paragraph{Data Generation}
	We generate data sets to test the quality of our surrogate models. To that end, we record the data
	of model-free optimization experiments. Here, optimization is performed with a quality diversity (QD) algorithm. 
	These algorithms are not only used to find good solutions but also are intended to find as many diverse and high-performing solutions as possible. We choose MAP-Elites~\cite{Mouret2015}, which builds up an archive of high-performing elites, one within each niche. 
	Here, niches are defined as cells in the grid shown in Figure~\ref{fig:maze}. 
	Parents are selected from the archive at random and their genes mutated with 5\% probability to form the next generation of controllers. These child controllers are tested and assigned the cell which corresponds to their end position. If the child arrived at that cell with a shorter path than the current occupant, it replaces the current occupant.

	\begin{figure}[hbp]
		\includegraphics{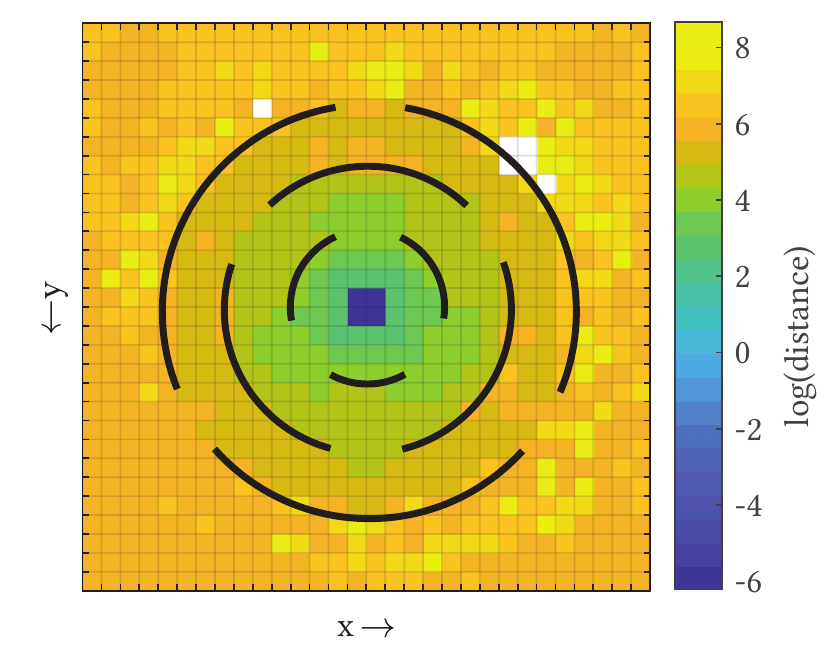}
		\caption{Distance map generated by MAP-Elites (lower distance equals higher fitness). Each niche in the map contains a robot controller that is optimized towards reaching that niche in the shortest path possible. }
		\label{fig:exampleMap}
	\end{figure}
	
	Figure~\ref{fig:exampleMap} shows an example distance map after 5000 generations, with almost each niche filled with a high-performing controller. The distance values grow the further they are from the center, which is to be expected. A number of controllers end up driving around the maze in circles, which explains the high distance values in some niches.
	
	\paragraph{Data Used for Modeling}
	Data generation was performed either with networks with 2 or 5 hidden neurons.
	We performed 20 replications, that is, we received data from 20 different QD runs for each experiment configuration, each with a different random number generator
	seed. This leads to 40 data sets (20 for each number of hidden neurons). 
	
	\begin{figure*}
		\includegraphics{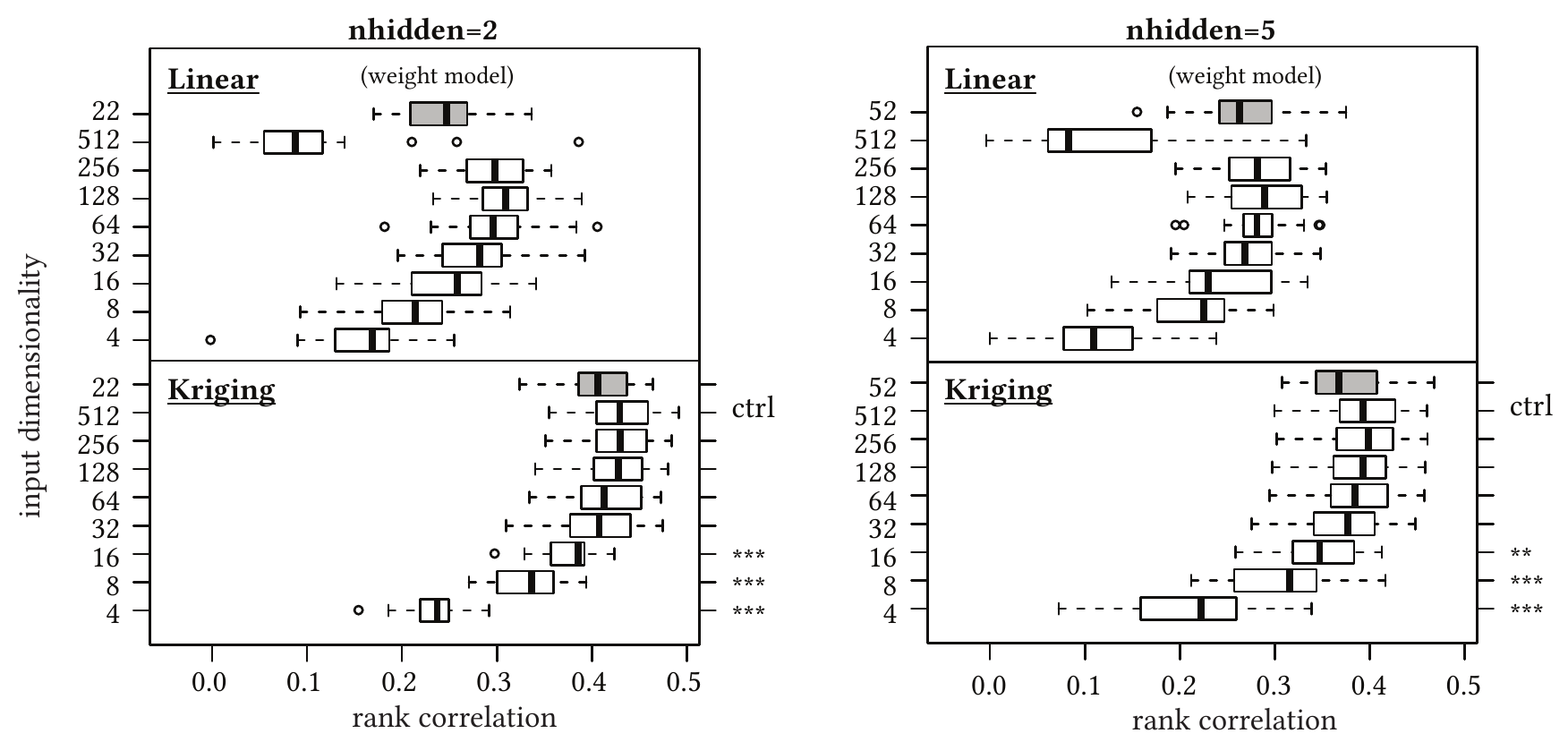}
		\caption{The model quality in terms of correlation (x-axis), for linear and Kriging models and different input spaces, and different numbers of hidden neurons (nhidden).
			Here, the numbers at the start of each y-axis label denotes the dimensionality of the
			input vector for the corresponding model. 
			Gray fill color indicates a model based on the weights or genotype, the white fill color indicates phenotypic models.
			The y-axis labels on the right-hand side indicate $p$-values from a statistical test that compares each of the Kriging models against the model marked with ctrl (*: $p<0.05$,**: $p<0.01$,***: $p<0.001$).
		}
		\label{fig:rescor}
	\end{figure*}
	
	Each data set consists of roughly 900 neural network controllers. For each of those controllers, we received nine different data subsets:  one with the weights, and eight with phenotypes of different sizes ($4, 8, \ldots, 512$).
	Note, that the phenotypes are derived from the two outputs of the networks, that is, if the network is fed with four input samples, we observed eight phenotype values. We can now describe each of the 900 controllers either by its weights, or by phenotype vectors (of different lengths).
	
	During modeling, these data sets are split as follows. 400 controllers are used to train a model, the remainder is used to test the model quality. Note that the observed values $y$ will be log-scaled before modeling, as the data contains strong outliers
	which might deteriorate the models.
	
	\paragraph{Quality Measures}
	To judge the quality of our models, we use Kendall's rank correlation coefficient~\cite{Kendall1990}.
	In contrast to Pearson correlation, this measure only considers ordinal correlation, i.e., the ranks of two compared sets of samples. Kendall correlation is therefore a good measure to estimate the accuracy of a model when used in rank-based evolutionary optimization methods.

	\paragraph{Kriging Model}
	We generate the Kriging model with the R-package CEGO~\cite{CEGOv2.3.0} as follows.
	For MLE, the optimization of the likelihood is performed via the locally biased variant of the Dividing Rectangles (DIRECT) algorithm~\cite{Gablonsky2001}. It is configured to stop after 2000 likelihood evaluations, or when a relative decrease in function values between iterations drops below $10^{-16}$. 
	The nugget effect (regularization) of the model is turned on, to potentially account for noise in the data or ill-conditioned kernel matrices.
	The model uses the Manhattan distance (see Section~\ref{sec:Kriging}).
	
	\paragraph{Comparison Baseline: Linear Model}
	We include a linear regression model in our experiments, as a comparison baseline for the Kriging model.
	Like the Kriging model, the linear model is trained  with the weight or the phenotype data.
	Since the generated data is potentially very high dimensional, we need some form of variable selection
	to generate reasonable models. We decided for a forward selection approach via the Aikake Information Criterion (AIC)~\cite{Venables2002}, starting from a model that only consists of an intercept.
	The most complex linear model may include main effects for all variables, but no interactions or higher order terms are considered.

	\subsection{Results and Discussion}
	Figure~\ref{fig:rescor} shows the Kendall correlation achieved by each of our models.
	Firstly, it can be observed that the Kriging model outperforms the linear model in most cases, as expected.
	Secondly, the variants based on phenotypic data are able to perform at least as well as the weight models, if the number of elements in the phenotype vector is at least 32 or more.  
	The larger phenotype vectors do not seem to yield much further improvement.

	We confirmed these observations by applying statistical tests for each number of hidden neurons.
	Firstly, we tested for the global presence of significant differences via the non-parametric Kruskal-Wallis rank-sum test~\cite{Kruskal1952}, which yielded $p$-values of less than $10^{-8}$ in both cases, indicating that differences are present. Afterwards,
	we performed Conover's non-parametric many-to-one comparison test~\cite{Conover1979}, comparing each of the Kriging models against a single model (control group).
	The chosen control group was the most complex model with phenotype data of dimensionality 512.
	The implementations of the employed tests were taken from the \texttt{stats}
	and the \texttt{PMCMRplus} \texttt{R} packages~\cite{RCoreTeam2018,Pohlert2018}:
	\texttt{kruskal.test} and \texttt{kwManyOneConoverTest}.
	The respective cases with indications for significant differences are marked on the right-hand side of each plot in Figure~\ref{fig:rescor}.
	The statistical test largely confirms the visual evaluation. 
	No evidence for differences is found between the control group and the model with the genotypic weight data. 
	Only models with phenotypic data of a dimensionality of 16 or less is deemed to be different from the control group.
	
	Importantly, the results suggest that we can use phenotypic surrogate models instead of those based on the genotype or weights.
	The phenotypic data is largely unaffected by the number of hidden neurons, and, hence, the number of weights. Where standard models would struggle to compare the weights of differently structure networks, a phenotypic comparison would still be possible.
	
	The baseline linear model shows some peculiar behavior. The model's performance drops off for models with phenotype vectors of more than 256 elements.
	This behavior can be largely explained with number of coefficients selected by the AIC forward selection procedure, as shown in Figure~\ref{fig:ncoef}. Clearly, the selection procedure will not select more than $n$ variables.
	The required number of variables seems to increase non-linearly with the dimensionality of the data.
	\begin{figure}
		\includegraphics{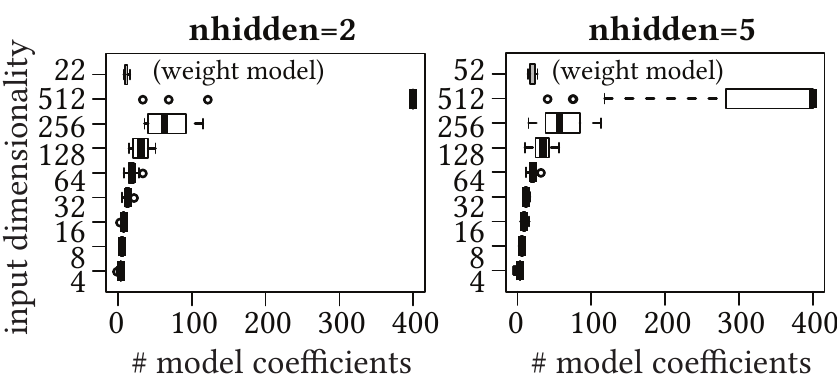}
		\caption{The number of linear model coefficients selected via forward selection based on AIC. Gray fill color indicates a model based on the weights or genotype, the remainder are based on phenotype data.}\label{fig:ncoef}
	\end{figure}
	
	Notably, the Kriging model does not show such a performance drop, and in fact performs quite well even for the very
	high dimensional phenotype vectors. 
	This may be counter-intuitive at first: Kriging is usually not suggested for high-dimensional data. We suggest two reasons for this: 
	Firstly, we use an isotropic model which avoids the complex optimization of fitting
	numerous kernel parameters ($\theta$).
	Secondly, there may be some sort of correlation in the observed phenotypes. 
	Increasing the number of samples used to generate the phenotype vector will increase the dimension,
	yet also increase the density in the sampled space. 
	In that sense, a new phenotype observation added to a large set of existing observations is likely
	to be quite similar to the existing observations. 
	Essentially, we assume that the latent dimensionality of the data is much lower.
	
	To verify this, we considered a Principle Component Analysis (PCA) of the input data (that is, excluding the dependent variable).
	For each of our data sets, we performed a PCA on the weight data, as well as on the phenotype data.
	In each case, we recorded the number of principal components required to explain 90\% of the variation
	in the data set. This number is shown in Figure~\ref{fig:pca}.
	\begin{figure}
		\includegraphics[width=1\linewidth]{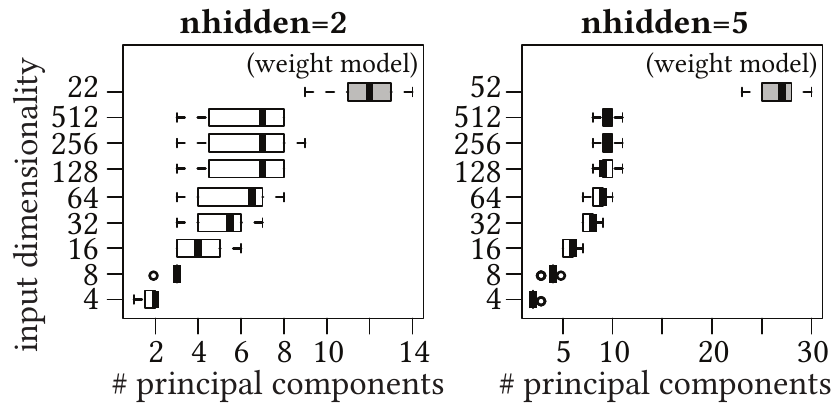}
		\caption{For each data set, the number of principal components required to explain 90\% of the variation in the data.
			This only concerns the respective input data of the surrogate models, the observed output (i.e., quality of the controller) is not considered here.
			Gray fill color indicates weight or genotype data, the remainder is based on phenotypic data.
		}\label{fig:pca}
	\end{figure}
	There are two interesting observations here. Firstly, the number of components levels off for the largest 
	phenotype vectors. The median stays at 7 (nhidden=2) and 9 (nhidden=5), despite data sets with several
	hundreds of variables. It seems that this confirms our assumption that the additional columns due to
	higher-dimensional phenotype vectors actually describe a much lower-dimensional, latent space.
	Secondly, we can see that the number of principal components for the weights are much larger.
	Yet, this does not coincide with better models based on the weight data.
	
	\section{Conclusions}
	In this work, we evaluated the use of phenotypic data of neural networks as a basis for surrogate modeling.
	We have shown that models based on phenotypic data can perform at least as well as those based on genotypic (weight) data. 
	This was true both for a baseline, linear model, and a non-linear Kriging model. 
	Our analysis further indicates that even high dimensional phenotypes with several hundreds of observations can yield sound Kriging models.
	A principal component analysis reveals that these high-dimensional data sets can be very well reproduced with only very few components. 
	A much larger number of components is required for the genotype data.
	
	This success of a phenotypic model is promising, since a model based on genotypes becomes infeasible
	if the compared networks have different structures or topologies, that is, in the context of evolutionary algorithms that can change the structure and size of the solution encoding, e.g. in surrogate-assisted neuroevolution.
	Measuring behavior of neural networks without using actual simulations not only seems to be possible, but also a practical way to compare networks. 
	
	Phenotypic distances can be used successfully as kernels to build surrogate models that predict the fitness of networks with varying sizes and topologies. 
	Whereas previous approaches to construct surrogate models of neural networks with non-uniform structure rely on the peculiarities of the evolutionary algorithm~\cite{gaier2018data}, our approach is independent of the optimization approach. In fact, a phenotypic distance approach to modeling is independent even of encoding: a neural network grown with NEAT, a fixed topology network optimized with particle swarm optimization, and a controller evolved with genetic programming could all share the same surrogate model.
	
	In future work, we plan to take the obvious next step: to actually use the developed models as surrogates in an optimization framework. In addition, we want to investigate the generation of phenotype vectors in more detail. As the PCA showed, as well as the diminishing returns for models with more phenotype samples, a lower-dimensional data set may suffice to produce good models. Creating better, more condensed phenotype samples with less redundant information is hence of interest for future work, to reduce the load of distance calculations.
	
	Being able to successfully model the performance of a robot controller by observing its behavior provides a computationally efficient and effective approach for surrogate modeling of varying-length representations. 
	We show that modeling the behavior of networks avoids some complexities that are caused by genotypic comparisons. 
	Surrogate-assisted optimization of non-uniform representations will allow a much more diverse set of solutions to be calculated with a limited number of real evaluations.

\bibliographystyle{ACM-Reference-Format}
\bibliography{phdDist} 

\end{document}